\documentclass{article}
\usepackage{spconf,amsmath,graphicx}
\usepackage{booktabs, multirow}
\usepackage{url}

\title{Stateful Conformer with Cache-based Inference for Streaming Automatic Speech Recognition}

\name{Vahid Noroozi$^1$, Somshubra Majumdar$^1$, Ankur Kumar$^2$\sthanks{He worked on this paper while doing an internship at NVIDIA.}, Jagadeesh Balam$^1$, Boris Ginsburg$^1$}
\address{$^1$NVIDIA\\ $^2$UCLA, Department of Computer Science}

\begin{document}
%\ninept
%
\maketitle
\begin{abstract}
In this paper, we propose an efficient and accurate streaming speech recognition model based on the FastConformer architecture. We adapted the FastConformer architecture for streaming applications through: (1) constraining both the look-ahead and past contexts in the encoder, and (2) introducing an activation caching mechanism to enable the non-autoregressive encoder to operate autoregressively during inference. The proposed model is thoughtfully designed in a way to eliminate the accuracy disparity between the train and inference time which is common for many streaming models. Furthermore, our proposed encoder works with various decoder configurations including Connectionist Temporal Classification (CTC) and RNN-Transducer (RNNT) decoders. Additionally, we introduced a hybrid CTC/RNNT architecture which utilizes a shared encoder with both a CTC and RNNT decoder to boost the accuracy and save computation.

We evaluate the proposed model on LibriSpeech dataset and a multi-domain large scale dataset and demonstrate that it can achieve better accuracy with lower latency and inference time compared to a conventional buffered streaming model baseline. We also showed that training a model with multiple latencies can achieve better accuracy than single latency models while it enables us to support multiple latencies with a single model. Our experiments also showed the hybrid architecture would not only speedup the convergence of the CTC decoder but also improves the accuracy of streaming models compared to single decoder models.

\end{abstract}
\begin{keywords}
Streaming ASR, FastConformer, Conformer, CTC, RNNT 
\end{keywords}
\section{Introduction}
\label{sec:intro}

Many of the traditional end-to-end streaming automatic speech recognition (ASR) models use auto-regressive RNN-based architectures \cite{he2019streaming} as we don't have access to all the future speech in streaming mode. Offline ASR models can potentially use the global context while streaming ASR models need to use a limited future context which degrades their accuracy compared to offline models. In some streaming approaches, offline models are being used for streaming which would be another source of accuracy degradation as there is inconsistency between offline training and streaming inference. The accuracy gap between streaming and offline models can be reduced by using large overlapping buffers where left and right contexts are added to each chunk of audio, however, this requires significant redundant computations for overlapping segments. 

In this paper, we propose an efficient and accurate streaming model based on the FastConformer \cite{fastconformer} architecture which is a more efficient variant of Conformer\cite{gulati2020conformer}. Our proposed approach would work with both the Conformer and FastConformer architectures but we performed our experiments with just FastConformer as it more than 2X faster than Conformer. We also introduced a hybrid CTC/RNNT architecture with two decoders of CTC \cite{graves2006connectionist} and RNNT \cite{graves2012sequence} with a shared encoder. It would not only saves computation as a single model is trained instead of two separate models but also improves the accuracy and convergence speed of the CTC decoder.

We propose a caching mechanism by converting the FastConformer's non-autoregressive encoder into an autoregressive recurrent model during inference using a cache for activations computed from previous timesteps. A cache stores the intermediate activations which are reused in future steps. The caching removes the requirement of any buffer or overlapping chunks which results in avoiding any unnecessary duplicate computations. It drastically reduces the computation cost when compared to traditional buffer-based methods. Note that the model is still trained efficiently in non-autoregressive mode, similar to offline models. 

The model has also limited right and left contexts during training to maintain consistent conditions during training and streaming inference. This consistency helps to reduce the accuracy gap between offline inference and streaming inference significantly. Additionally, as the changes are limited to the encoder architecture, the proposed approach works for both FastConformer-CTC and FastConformer-Transducer (FastConformer-T) models. We evaluate the proposed streaming model on the LibriSpeech dataset and a large multi-domain dataset and show that it outperforms the buffered streaming approaches in terms of accuracy, latency, and inference time. We also study the effect of the right context on the trade-off between latency and accuracy. In another experiment, we would evaluate a model trained with multiple latencies which can support multiple latencies in a single model. In our experiments, we show that it can achieve better accuracy than models trained with single latency. Additionally we show that our hybrid architecture can achieve better accuracy compared to single decoder models with less compute. All the code and models used in the paper including the training and inference scripts are open-sourced in NeMo \cite{kuchaiev2019nemo} toolkit\footnote{https://github.com/NVIDIA/NeMo}.

\section{Related Works}
There are a number of approaches that use limited future context in streaming models. The time-restricted methods in \cite{zhang2020transformer, moritz2020streaming} use masking in each layer to allow a limited look-ahead for each output token. However, these methods are not computationally efficient since the computations for look-ahead tokens are discarded and they need to be recomputed again for future steps. Another approach is based on splitting the input audio into several chunks. Each output token corresponding to a chunk has access to all input tokens in the current chunk as well as a limited number of previous chunks. This approach is more efficient and accurate than the time-restricted method \cite{chen2021developing}.  

Some memory-based approaches \cite{tsunoo2021streaming, wu2020streaming,inaguma2020enhancing} use contextual memory to summarize older chunks into a vector to be used in the subsequent chunks. For example, a streaming Transformer model  in \cite{tsunoo2021streaming} with an attention-based encoder/decoder (DEA) architecture uses a context embedding to maintain some memory state between consecutive chunks. Generally, these techniques are computationally efficient for inference, but they usually break the parallel nature of training, resulting in less robust and efficient training. 

There exist a number of previous works that adopted Conformer for streaming ASR \cite{li2021better,yu2020dual,yao2021wenet}. In \cite{li2021better,yu2020dual}, the authors have developed a unified model which can work in both streaming and non-streaming modes. Yao et al. \cite{yao2021wenet} proposed a streaming Conformer that uses a Transformer decoder. Their model supports dynamic look-ahead by training the model with different look-ahead sizes.

\section{Cache-aware Streaming FastConformer}
\label{sec:proposed}
In our proposed cache-aware streaming FastConformer, the left and right contexts of each audio step are controlled and limited. It enables us to have consistent behavior during both training and inference. The proposed model is trained in an efficient non-autoregressive manner, but inference is done in an autoregressive, recurrent way. 

The original FastConformer encoder consists of self-attention, convolutions, and linear layers. The linear layers and 1D convolutions with kernel size of one do not need any context because their outputs in each step are just dependent on that step. However, self-attention and convolutions with kernel size larger than one need context, and we need to limit the context for these specific layers to control the context size of the whole model.

\subsection{Model training} \label{sec:model_training}
We modify the FastConformer model as follows to adapt it for the streaming scenario. We avoid using normalization in the mel-spectrogram feature extraction step as the normalization procedure need to use of some statistics which depend on the entire input audio. We make all the convolution layers including those in the downsampling layers fully causal. For this purpose, we use padding of size $k-1$ to the left of the input sequence where $k$ is the convolution kernel size, and zero padding for the right side. From now onwards, we will drop the ``downsampled" prefix and simply refer to the ``downsampled input" as ``input". We replace all the batch normalization \cite{ioffe2015batch} layers with layer normalization \cite{ba2016layer} as the former computes mean and variance statistics from the entire input sequence whereas the latter normalizes each step of the input sequence independently. 

There are three approaches to limit the size of the context for self-attention layers:

\textbf{Zero look-ahead}: Zero look-ahead means each step in the sequence has access only to previous tokens (either all past tokens or only a subset of them). This is crucial for low-latency applications. Therefore, all modules need to be causal including the self-attention layers. We use masking to ignore the contribution of all future tokens in the attention score computation. It results in a small latency and inference time but lower prediction accuracy.

\begin{figure}
    \centering
    \includegraphics[scale=0.35]{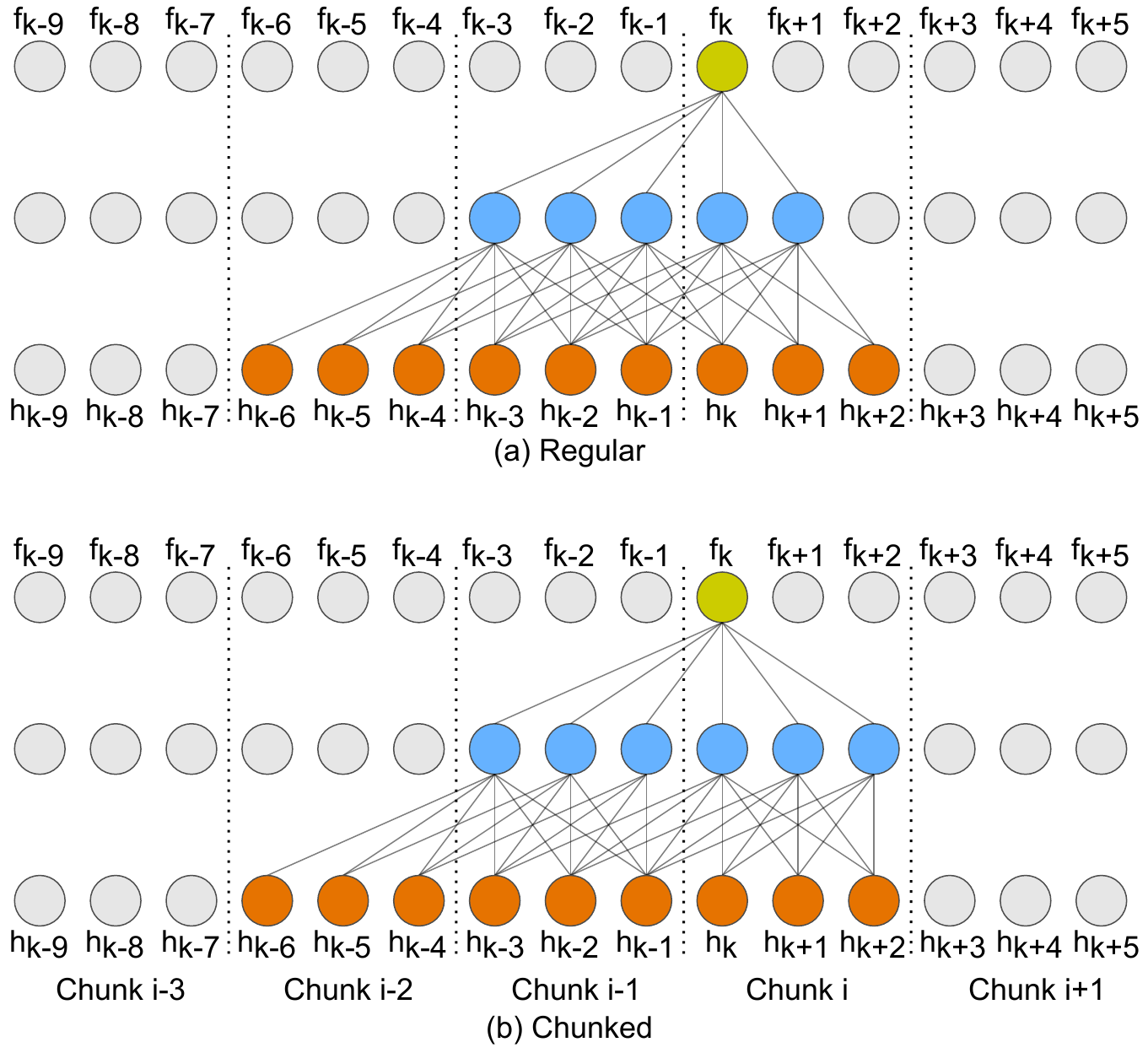}
    \caption{Diagram of how context gets extended with multi-layer self-attention layers in regular look-ahead vs chunk-aware. Dependency on future frames increases for regular look-ahead in self-attention layers as we go deep in the network whereas it remains the same for chunk-aware approach.}
    \label{fig:regular_vs_chunk_lookahead}
\end{figure}

\textbf{Regular look-ahead}: It has been shown that having access to some future time steps, i.e. limited look-ahead, can significantly improve the accuracy of an ASR model \cite{chen2021developing}.
The simplest approach is to allow a small look-ahead in each self-attention layer \cite{moritz2020streaming,zhang2020transformer}. Since layers are stacked, the effective look-ahead gets multiplied by the number of self-attention layers as we move deeper in the network as shown in Figure \ref{fig:regular_vs_chunk_lookahead}(a). For example, in a model with $N$ self-attention layers, where each one has a look-ahead of $M$, the effective look-ahead of each output token over the input sequence is $M \times N$. The effective look-ahead directly impacts the final latency since the model needs to wait for $M\times N$ time steps before it can make any prediction.
The past context in this approach can be any number of tokens. But allowing larger past context will increase the computation time for each streaming step.

\textbf{Chunk-aware look-ahead}: There are two disadvantages to the regular look-ahead. First, the effective overall look-ahead depends on the number of layers having non-zero look-ahead. Thus, the latency can be significant if we use look-ahead in each layer. Even for a reasonably large latency budget, we can only use a small look-ahead size in each layer (which we denote $M$). For example, for a model with $17$ layers, a frame rate of $10$ms, and subsampling factor of $4$, choosing look-ahead size $M=2$ results in a latency of $10\times4\times17\times2=1360$ milliseconds. Therefore, $M$ cannot be much larger for practical applications \cite{Kim2021MultimodeTT}.

The other disadvantage of regular look-ahead is the unnecessary re-computation of some tokens during the streaming inference. For example, to compute $f_k$ in figure \ref{fig:regular_vs_chunk_lookahead}(a), the self-attention operation is applied on blue tokens with a query size of $M+1$, 1 for current timestep $k$ and $M$ for future tokens (more details in section \ref{sec:inference_with_caching}). This generates the gray-shaded token $f_{k+1}$ along with the desired output $f_k$ shown in yellow. But we drop $f_{k+1}$ generated in this step as it is not correct due to its dependency on $h_{k+3}$, which is not available yet. Therefore, we need to recompute $f_{k+1}$ along with other such tokens across different layers.

Chunk-aware look-ahead \cite{chen2021developing,wang2020low,tian2020synchronous} addresses both the above issues. It splits the input audio into chunks of size $C$. 
Tokens in a chunk have access to all other tokens in the same chunk as well as those belonging to a limited number of previous chunks. In contrast with the effective look-ahead growing with depth in regular look-ahead, there is no such dependency in chunk-aware look-ahead. 
Due to chunking, the output predictions of the encoder for all the tokens in each chunk will be valid and there is no need to recompute any activation for the future tokens. This results in zero duplication in the compute and makes the inference efficient. While the look-ahead of each token is the same for regular look-ahead by construction, it varies in the range $\big[0, C-1\big]$ for the chunk-aware case. The leftmost token in a chunk has the maximum look-ahead with access to all the future tokens in the chunk whereas the last token has the least look-ahead with access to zero future tokens. The average look-ahead for any token in chunk-aware look-ahead is larger than the regular look-ahead which leads to better accuracy with the same latency budget.

\subsection{Hybrid architecture}
\begin{figure}
    \centering
    \includegraphics[scale=0.5]{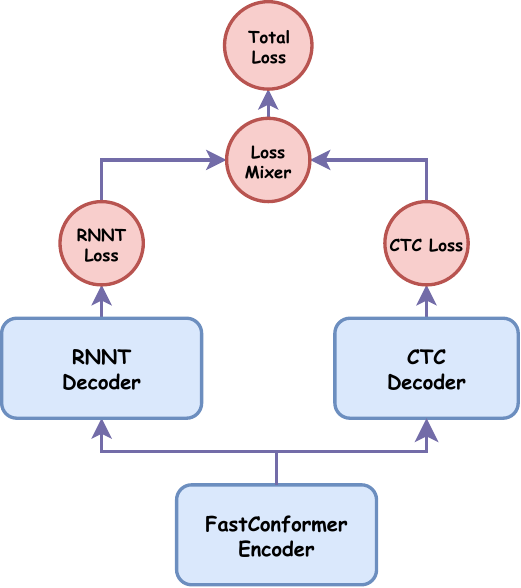}
    \caption{Architecture of the hybrid CTC/RNNT model.}
    \label{fig:hybrid_arch}
\end{figure}

We used a hybrid architecture which uses two decoders, one CTC decoder and one RNNT decoder train our models. Both decoders share a single encoder. The architecture of our hybrid model is shown in Figure \ref{fig:hybrid_arch}. After training is done, any of the two decoders can be used for inference. The hybrid architecture has the following advantages over single decoder models: 1) no need to train two separate models and saves significant compute in our experiments as we did all of experiments for both the CTC and RNNT, 2) speeds up the convergence of the CTC decoder significantly which is generally slower than RNNT decoders, and 3) improves the accuracy of both the decoders likely due to the joint training. During the training the losses of the CTC decoder ($l_{ctc}$) and RNNT decoder ($l_{rnnt}$) are mixed with a weighted summation as the following:

\begin{align*}
  & l_{total} = \alpha * l_{ctc} + l_{rnnt}
\end{align*}

where $l_{total}$ is the total loss to get optimized, and $\alpha$ is the hyperparameter to control the balance between these two losses.

\subsection{Inference with caching} \label{sec:inference_with_caching}
In streaming inference, we process the input in chunks. Using a larger chunk size results in higher latency but requires fewer calls to the forward pass through the model. We use chunk size $C=M+1$, where $M$ is the look-ahead. However, the chunks are overlapping with stride of $1$ in regular look-ahead compared to stride of $M+1$ with no overlap for chunk-aware look-ahead. The straightforward approach to process chunks is to pass each chunk along with the effective past context. However, this approach is very inefficient as there is a huge overlap in the computation of past context. We propose a caching approach to avoid these recomputations and have zero duplication in streaming inference. Normalization, feedforward, and pointwise convolution layers do not need caching as they do not require any context.
However, self-attention and depth-wise convolution with a kernel size greater than $1$ do depend on past context. Therefore, caching intermediate activations from the processing of previous chunks can lead to a more efficient inference. 

\begin{figure}
    \centering
    \includegraphics[scale=0.35]{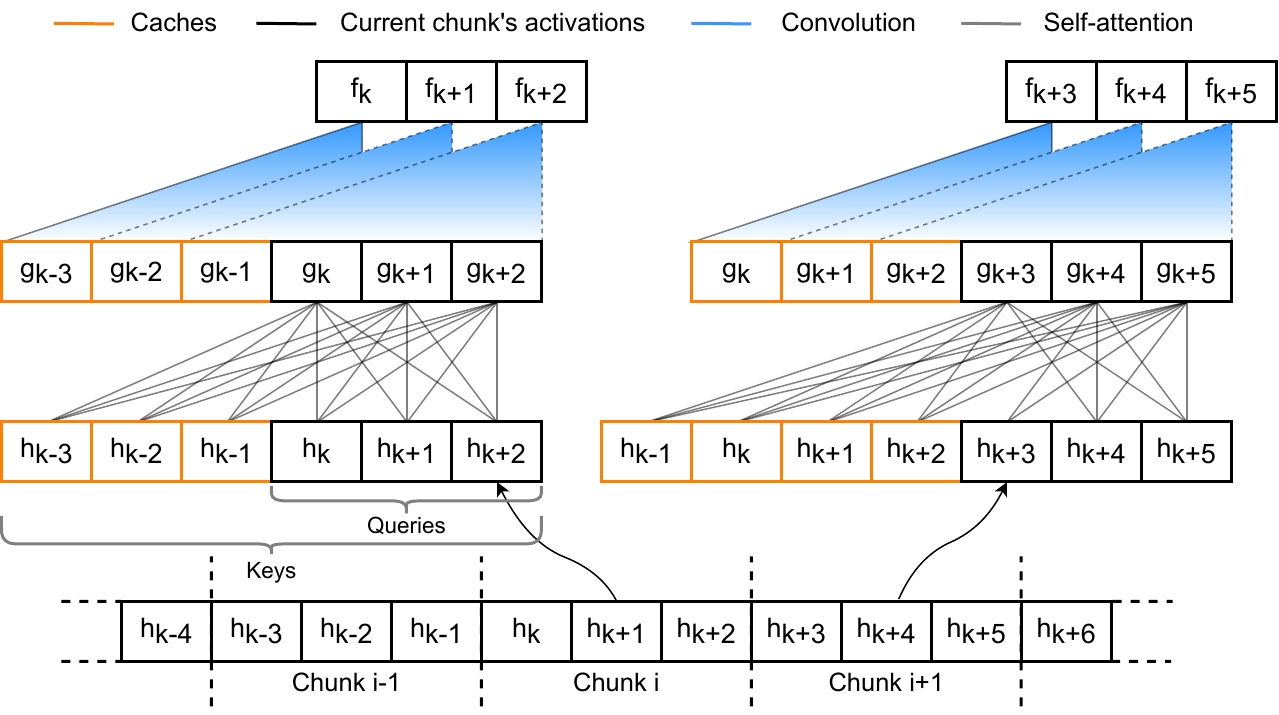}
    \caption{Caching schema of self-attention and convolution layers for consecutive chunks.}
    \label{fig:caching}
\end{figure}

For each causal 1D depthwise convolution with kernel size $K$, we use a cache of size $C_{conv} = K-1$. This cache contains the activations of the last $C_{conv}$ steps from the previous chunks. Initially, the cache is filled with zeros for the first chunk. It gets updated at each streaming step as shown in Figure \ref{fig:caching}. The cache is filled with the $g_{k-3}, g_{k-2}, g_{k-1}$ outputs of the layer below from the previous streaming step. In the current step, outputs $g_k, g_{k+1}, g_{k+2}$ from the layer below would be used to overwrite the previous values in that part of the cache. The updated cache therefore contains $g_{k+1}, g_{k+2}, g_{k+3}$ to be used in the next streaming step. Given a batch size of $B$ and a model with $L$ depth-wise convolutions layers, each having hidden size of $D$, we require a cache matrix of size $L\times B \times  D\times C_{conv}$. Each layer updates the cache matrix by storing the necessary activations in each streaming step.

Unlike the fixed-length cache for convolution layers, the cache size for a self-attention layer grows from zero up to the past context size. For self-attention layers with left context of $L_c$, the cache is empty in the first streaming step. With every streaming step, chunk size number of activations from the input to self-attention layer is added to the cache and any extra old values are dropped. Eventually, the cache grows to its full size and contains the last $L_c$ activations only. For example, in figure \ref{fig:caching}, the cache for self-attention contains only three values $h_{k-3}, h_{k-2}, h_{k-1}$ since initially the cache was empty and got updated with chunk size of three elements from previous streaming step. At the end of this step, $h_k, h_{k+1}, h_{k+2}$ are added to cache which would make cache size 6. Therefore, the two oldest values $h_{k-3}, h_{k-2}$ are dropped to maintain maximum cache size of $L_c$, here four, to be used in the next streaming step as shown on the right. For a batch size of $B$, a model with $L$  self-attention layers, each having hidden size of $D$, requires a cache matrix of size $L \times B\times C_{mha} \times D$ where $0 \le C_{mha} \le L_c$. 

The downsampling module uses striding convolutions and can also benefit from caching. However, due to a small kernel size (typically 3), it would be a small cache and can be ignored. Instead, we can simply concatenate the last $log(D_r)*2+1$ mel-spectrogram feature frames to each chunk where $D_r$ is the downsampling rate. The decoder of FastConformer-CTC is stateless while the RNN-T decoder consists of RNN layers with states. Therefore, for FastConformer-T, all the hidden states of RNN layers need to be stored after each streaming step. In the next step, these cached states are used to initialize all RNN layers. By maintaining such caches, the prediction of the network would be exactly the same as when the entire audio is processed in a single step.

\begin{table}
\small
    \centering
    \begin{tabular}{ccccc}
        \toprule
        %\cmidrule(lr){4}
            & Inference & Look-ahead & WER, & Avg Latency,\\
         Decoder & Mode & Method &\% &  ms \\
        \midrule
        \multirow{4}{*}{CTC} & Offline & Full context & 5.7 & - \\
        & Buffered & Buffering & 8.0 & 1500 \\
        %\cmidrule(lr){2-4}
         & \multirow{3}{*}{Cache-aware} & Zero & 10.6  & 0\\
         &  & Regular & 7.9 & 1360 \\
         &  & Chunk-aware & 7.1 & 1360\\
        \midrule
        \multirow{4}{*}{RNNT} & Offline & Full context & 5.0 & -\\
        & Buffered & Buffering & 11.3 & 2000\\
        %\cmidrule(lr){2-4}
        & \multirow{3}{*}{Cache-aware} & Zero & 9.5 & 0\\
        &  & Regular & 7.1 & 1360\\
        &  & Chunk-aware  & 6.3 & 1360  \\
        \bottomrule
    \end{tabular}
   \caption{The accuracy and average of latency for different streaming FastConformer models trained on LibriSpeech and evaluated on test-other set.
}
    \label{tab:streaming_vs_offline}
\end{table}

\begin{table*}[t]
\small
    \centering
    \begin{tabular}{ccccccccc}
         \toprule
         Decoder   & Architecture Type &  Approach    & 0ms  & 40ms& 240ms & 520ms & 680ms & 1360ms\\
         \midrule
         CTC & Hybrid & Regular       & 10.6 & -    & -   & -  & 8.3   &  7.9\\
             & Hybrid & Chunk-aware   & 10.6 & 10.1 & 8.8 & 8.4  & 8.0 &  7.1\\
             & Non-hybrid & Chunk-aware   & 10.8 & 10.3 & 8.9 & -  & 8.1 &  -\\
          \midrule
         RNNT& Hybrid & Regular   & 9.5  & -    & -   & -  & 7.6   &  7.1\\
              &  Hybrid & Chunk-aware      & 9.5  & 9.0  & 7.8 & 7.5  & 7.3 &  6.3\\
              &  Non-hybrid & Chunk-aware  & 9.4  & 8.9  & 8.3 & -  & 7.6 &  -\\
         \bottomrule
    \end{tabular}
    \caption{The accuracy (WER\%) of cache-aware streaming FastConformer models with different latencies and look-ahead approaches, evaluated on LibriSpeech test-other set. Not all latencies are feasible for regular look-ahead approach.}
    \label{tab:look_ahead_size_ls}
\end{table*}

\begin{table*}[t]
\fontsize{7pt}{7pt} \selectfont

    \centering
    \begin{tabular}{ccccccccccccc}
         \toprule
               &          & Avg. & LS \cite{panayotov2015librispeech} & LS \cite{panayotov2015librispeech} & SPGISpeech & Earnings22 & GigaSpeech & Tedlium & MCV & Voxpopuli & AMI & \\
         Model & Decoder  & Latency  & test-other & test-clean &  \cite{o2021spgispeech} &  \cite{del2022earnings}&  \cite{chen2021gigaspeech} &  \cite{rousseau2012ted}&  \cite{ardila2019common} &  \cite{wang2021voxpopuli}&  \cite{kraaij2005ami} & Averaged\\
         \midrule
          Cache-aware&CTC      & 40  & 7.9  & 3.4   & 7.1  & 22.5    & 16.7 & 7.1& 15.8& 9.9& 29.3& 13.3\\
          Cache-aware&       & 240 & 7.3  & 3.4   &6.6  & 22.2    &  15.8& 6.5& 15.1& 8.8& 27.3& 12.6\\
          Cache-aware &     & 520 & 6.2  & 2.6   & 6.1  & 21.2    &  14.3& 5.8& 13.6& 7.8& 23.3& 11.2\\
          Buffered&     & 2000 & 7.7  & 3.6 & 6.8  & 20.1 &  14.0       & 5.5 & 15.9 & 8.8 & 25.3& 12.0\\
         \midrule

         Cache-aware& RNNT     & 40  & 6.4  & 2.6 & 6.1  & 21.0 &  15.2& 6.2& 13.8& 8.2& 29.1& 12.1\\
          Cache-aware&     & 240 & 5.9  & 2.5 & 5.7  & 20.8 &  14.2& 5.5& 13.0& 7.6& 26.8& 11.3\\
          Cache-aware&     & 520 & 5.4  & 2.2 & 5.5  & 20.0 &  13.6& 5.4& 11.9& 7.1& 24.2& 10.6\\
          Buffered &     & 2000 & 9.4  & 4.8   & 8.8  & 23.8    &  16.4& 7.0&1 7.7 & 10.8& 36.0& 15.0\\
         \bottomrule
    \end{tabular}
    \caption{The accuracy (WER\%) of cache-aware and buffered streaming FastConformer with different look-ahead sizes and decoders on different benchmarks. All models are trained on NeMo ASRSET 3.0.}
    \label{tab:look_ahead_size_nemo}
\end{table*}

\begin{table}
\small
    \centering
    \begin{tabular}{cccccc}
         \toprule
         Model            & Decoder & 40ms& 240ms & 520ms \\
         \midrule
         Single-Lookahead & CTC & 7.9 & 7.3 & 6.2 \\
         Multi-Lookahead   &    & 7.6 & 6.5 & 6.0  \\
         \midrule
         Single-Lookahead & RNNT & 6.4 & 5.9 & 5.4  \\
         Multi-Lookahead   &     & 6.2 & 5.5 & 5.2   \\
         \bottomrule
    \end{tabular}
    \caption{The comparison between single look-ahead models vs a multi-lookahead model trained on NeMo ASRSET 3.0 for different latencies. Accuracies (WER\%) are reported on test-other set of LibriSpeech.}
    \label{tab:single_vs_multi}
\end{table}

\section{Experiments}
\label{sec:exps}

We evaluated our proposed streaming approach with hybrid architecture of FastConformer \cite{fastconformer}. All the results are reported for both the the CTC and RNNT decoders which are denoted as FastConformer-CTC and FastConformer-T respectively. The parameter $\alpha$ of the hybrid loss is set to $0.3$ as it showed the best performance in our experiments. We performed all the experiments on the models which have $\approx114$M parameters. We followed the same configuration used in \cite{fastconformer}. Experiments are done on two datasets: 1) LibriSpeech (LS) with speed perturbation of 10\% \cite{panayotov2015librispeech}, and 2) NeMo ASRSET 3.0. NeMo ASRSET is a large multi-domain dataset which is a collection of some publicly available speech datasets with total size of 26K hours. 

All models are trained for at least 200 epochs with effective batch size of 2048 for LibriSpeech and 4096 for NeMo ASRSET 3.0. SentencePiece \cite{kudo2018sentencepiece} with byte pair encoding (BPE) is used as the tokenizer with vocab size of 1024, and they are trained on the train set of each training dataset. We trained the models with AdamW optimizer \cite{loshchilov2017decoupled} with weight decay of $0.001$ and Noam scheduler \cite{vaswani2017attention} with coefficients of $5.0$. We used checkpoint averaging of the best five checkpoints based on the WER of the validation sets to get the final models. Mixed precision training with FP16 \cite{micikevicius2018mixed} is used for most of the experiments to speed up the training process. All the average latencies in this paper are referring to algorithmic latency induced by the encoder (EIL) introduced in \cite{shi2021emformer}. It is calculated as the average time needed for each word to get predicted by the model while ignoring the inference time of the neural network.

We used FastEmit \cite{yu2021fastemit} for the RNNT loss with $\lambda$ of $0.005$ to prevent the model from delaying the predictions. FastEmit showed to be very effective and crucial to improve the accuracy of the streaming models for both the RNNT and CTC decoders. This positive cross-decoder effect on the CTC decoder is another advantage of the hybrid architecture.

\subsection{Streaming vs offline models}

In this experiment, we compare different cache-aware streaming models with offline models and buffered streaming. All models are trained on LS and the results of the evaluations on the test-other set of the LS are reported in Table \ref{tab:streaming_vs_offline}. The offline models are trained with unlimited context over the entire audio. We evaluated and reported the performance of these models in both full context as well as buffered streaming mode. We use the buffered streaming solution as a baseline which can be used for streaming inference with models trained in full-context (offline) mode. In this approach, the input is passed chunk-by-chunk but in order to get reasonable results at the borders, we add some of the past and future audios as context to each chunk. The total audio including the chunk and its contexts is stored in a buffer. The contexts would result in re-computation and waste of compute. In the experiments for buffered streaming, we used a chunk size of $1$ second and $2$ seconds for the CTC and RNNT respectively with buffer sizes of 4 seconds. 

The results for the regular and chunk-aware streaming models are selected from the models with average latency of 1360ms. Cache-aware models show significantly better accuracy with lower latency while using less computation compared with the buffered approach. While some contexts are added to each chunk in buffered streaming, not that much duplication is needed for chunk-aware streaming models. It makes the cache-aware streaming models significantly faster than buffered streaming models. This speed gap can be significantly higher with a larger buffer size or smaller chunk size. 

As it can be seen, the accuracy of the buffered streaming for RNNT models is not as good as the CTC decoders while they use even larger chunk size. Additionally, in our experiments the performance of the buffered RNNT was not robust to the buffer and chunk size parameters, while cache-aware models were more robust and showed better accuracy with lower latency.

Moreover, our streaming models show smaller accuracy degradation from the offline model compared to buffered streaming. The accuracy of our streaming model is exactly the same when evaluated in offline and streaming modes as training and evaluation have the same limited contexts. There is inconsistency between the contexts available during training and inference for the buffered approach. Due to the caching mechanism, the total computation for offline inference and streaming inference is also the same for chunk-aware approach.

\subsection{Effect of look-ahead size on accuracy}
We evaluated the effect of different look-ahead sizes on the accuracy of the proposed streaming models on LibriSpeech dataset. The WERs on test-other set of LS for six different lengths of look-ahead are shown in Table \ref{tab:look_ahead_size_ls} for regular and chunk-aware approaches. One of the disadvantages of the regular look-ahead over the chunk-aware is that not any look-ahead size is feasible with regular look-ahead. For FastConformer models which has 8X downsampling and window shift of the mel spectogram input is 10ms, even one token of look-ahead would translate into $8*10*17=1360ms$ of look-ahead considering all the evaluated models have 17 layers.

Results show that chunk-aware look-ahead are better than regular look-ahead in term of accuracy with the same latency. Additionally, it can be seen that larger look-ahead significantly improves the accuracy of both approaches, which shows the importance of look-ahead for better accuracy with the sacrifice of latency. The average of latency for each case is half of the look-ahead size. 
In the same Table \ref{tab:look_ahead_size_ls}, we also reported the accuracy of the same models with chunk-aware approach trained with non-hybrid architecture to show the effectiveness of the hybrid architecture for streaming models. As it can be seen, the hybrid variants demonstrate better accuracy compared to non-hybrid ones.

\subsection{Large scale multi-domain training}
To evaluate the effectiveness of our proposed approach, we evaluated the chunk-aware model on a large multi-domain dataset (NeMo ASRSET 3.0). More detail on this dataset can be found in \cite{fastconformer_large_streaming_multi}.

The accuracy of both the cache-aware FastConformer-CTC and FastConformer-T evaluated on a collection of evaluation sets are reported in Table \ref{tab:look_ahead_size_nemo}. As expected results are similar to the experiments on the LibriSpeech, higher latency would result in higher accuracy, and RNNT-based models are better than their equivalent CTC. As it can be seen, the cache-aware streaming models outperform the buffered streaming models on all benchmarks.

\subsection{Multiple look-ahead training}
One of the disadvantages of the cache-aware streaming comparing to buffered streaming is that each model is trained for a specific latency and supporting multiple latencies need training of multiple models. In order to address this shortcoming, we proposed to train the streaming model with multiple latencies. For each batch on each GPU, we randomly select a chunk size and it makes the model to support different latencies. 
To evaluate the proposed approach, we trained a chunk-aware model with multiple latencies and compared the averaged accuracy on all benchmarks to models trained with single latency. The benchmarks are the same as the ones used in Table \ref{tab:look_ahead_size_nemo}. The multi-lookahead model may need more steps to achieve the same accuracy as training a single latency model. The results are reported in Table \ref{tab:single_vs_multi} for both the CTC and RNNT decoders. The multi-lookahead model even shows better accuracy than single lookahead models while just one model is trained for multiple latencies. The training on multiple look-aheads have helped the model to become more robust and even achieve better accuracy in some cases.

\section{Conclusion}
We proposed a streaming ASR model based on FastConformer where the non-autoregressive encoder is converted into an autoregressive recurrent model during inference. It is done by using an activation cache to keep the intermediate activations that are reused in future steps. The caching drastically reduces computation cost when compared to traditional buffer-based methods while the model is still trained in non-autoregressive mode. We evaluated our proposed model on LibriSpeech and a large multi-domain dataset and showed that the proposed model outperforms buffered streaming in terms of accuracy, inference time, and latency. We also introduced a hybrid CTC/RNNT architecture to train the streaming models which not only saved compute but also improved the accuracy. Additionally our experiments showed that a model trained with multiple latencies can achieve even better accuracy than models trained with single latency.

% \vfill
% \pagebreak

\bibliographystyle{IEEEbib}
\bibliography{refs}

\end{document}